\pdfoutput=1
\documentclass{article}

\PassOptionsToPackage{numbers, compress}{natbib}
\usepackage[preprint]{neurips_2023}
\usepackage{graphicx} 
\usepackage{tabularx}
\usepackage[ruled,vlined]{algorithm2e}
\usepackage{algpseudocode}
\usepackage{amsthm}
\usepackage{booktabs}
\usepackage{adjustbox}
\usepackage{caption}
\usepackage{subcaption}

\usepackage{xcolor}
\usepackage{kotex}% menu에서 texlive version을 2022에서 2021로 변경한 후 사용할 수 있었음. 

\usepackage{xcolor,colortbl}
\newcommand\hb{\rowcolor{orange!10}}

\usepackage[utf8]{inputenc} % allow utf-8 input
\usepackage[T1]{fontenc}    % use 8-bit T1 fonts
\usepackage{url}            % simple URL typesetting
\usepackage{booktabs}       % professional-quality tables
\usepackage{amsfonts}       % blackboard math symbols
\usepackage{nicefrac}       % compact symbols for 1/2, etc.
\usepackage{microtype}      % microtypography
\usepackage{xcolor}         % colors
\usepackage{amsmath}
\usepackage[colorlinks, linkcolor=blue]{hyperref}

\title{Efficient Storage of Fine-Tuned Models via Low-Rank Approximation of Weight Residuals}

\author{
  Simo Ryu\thanks{Equal contribution} \\
  KAIST \\
  \texttt{\small{clonofsimo@gmail.com}} \\
  \And
  Seunghyun Seo\footnotemark[1] \\
  NAVER Cloud \\
  \texttt{\small{real.seunghyun.seo@navercorp.com}} \\
  \And
  Jaejun Yoo \\
  UNIST \\
  \texttt{\small{jaejun.yoo@unist.ac.kr}} \\
}

\begin{document}

\maketitle
\newcommand{\rsm}[1]{\textcolor{blue}{SM: #1}}
\newcommand{\kmj}[1]{\textcolor{red}{MJ: #1}}
\newcommand{\ssh}[1]{\textcolor{blue}{SH: #1}}
\newcommand{\yjj}[1]{\textcolor{green}{JJ: #1}}

\newcommand{\todo}[1]{\textcolor{red}{TODO: #1}}

\newcommand{\placeholder}[0]{\textcolor{red}{Lorem ipsum dolor sit amet, consectetur adipiscing elit, sed do eiusmod tempor incididunt ut labore et dolore magna aliqua. Sit amet porttitor eget dolor morbi. Aliquet nibh praesent tristique magna sit amet. Amet dictum sit amet justo donec. Orci sagittis eu volutpat odio facilisis mauris. Nibh sit amet commodo nulla. Etiam non quam lacus suspendisse. Hac habitasse platea dictumst quisque. Ultrices sagittis orci a scelerisque purus semper eget. Vestibulum morbi blandit cursus risus at ultrices mi tempus imperdiet.}}
\begin{abstract}

In this paper, we present an efficient method for storing fine-tuned models by leveraging the low-rank properties of weight residuals. Our key observation is that weight residuals in large overparameterized models exhibit even stronger low-rank characteristics. Based on this insight, we propose Efficient Residual Encoding (ERE), a novel approach that achieves efficient storage of fine-tuned model weights by approximating the low-rank weight residuals. Furthermore, we analyze the robustness of weight residuals and push the limit of storage efficiency by utilizing additional quantization and layer-wise rank allocation.
Our experimental results demonstrate that our method significantly reduces memory footprint while preserving performance in various tasks and modalities. We release our code. \footnote{Our code is available at:
\url{https://github.com/cloneofsimo/ere}}

\end{abstract}

\section{Introduction}

Fine-tuning large-scale pre-trained models has become a widely adopted technique in tasks such as language modeling~\cite{bert,bart, howard2018universal}, speech recognition~\cite{baevski2020wav2vec} and image generation~\cite{ruiz2023dreambooth, ftgan1, ftgan2}. 
It has proven to be effective in achieving state-of-the-art performance on diverse tasks. 
However, as the number of model parameters continues to grow exponentially, reaching billions, storing fully fine-tuned parameters for each specific task becomes impractical and inefficient. 
To tackle this issue, Parameter Efficient Fine-Tuning (PEFT) approaches~\cite{karimi2021compacter,lora,li2021prefix, prompt_tuning, bitfit,adapter} have been introduced, aiming to tune only a subset of the original parameters and preserving the tuned parameters for individual tasks.

While PEFT methods are efficient and show promising results that are competitive with full fine-tuned models, they often rely on complex handcrafted modules or fixed-dimensional hyperparameters for each model layer.
For example, methods like Compacter \cite{karimi2021compacter}, LoRA \cite{lora}, and prefix-tuning \cite{li2021prefix} introduce additional complexity by requiring specific ranks or token prefixes, and achieving optimal performance with these methods often necessitates extensive hyperparameter search. Recent research has also shown that even with thorough incorporation of extensive model/task/resource-specific priors, PEFT struggles to outperform full fine-tuning in low-to-medium resource scenarios \cite{peftvsft2} under compute budget. Downside of PEFT becomes more pronounced as these methods are not compatible with existing techniques that are done on fine-tuning of full weights \cite{ewc, vcl, synint, aljundi2018memory, fedavg, robustft, modelsoup, matena2022merging}

For these reasons, one might conclude to use full fine tuning for their downstream fine-tuning.
If so, \textbf{can we still resolve the problem of inefficient memory footprint in a full fine-tuning setup?} To tackle this problem, we propose a different approach in this paper focusing on weight residuals: the differences between fine-tuned models and their base counterparts.
Remarkably, we have observed that weight residuals of overparameterized models, already acknowledged for their low-rank nature \cite{aghajanyan2021intrinsic, measureintrin, huhlow, heavytailed}, exhibit even more pronounced low-rank characteristics. 
Furthermore, these individual residual parameters demonstrate robustness, indicating that their contributions to downstream data shift are highly redundant.

\begin{figure}
  \centering
    \includegraphics[width=0.75\textwidth]{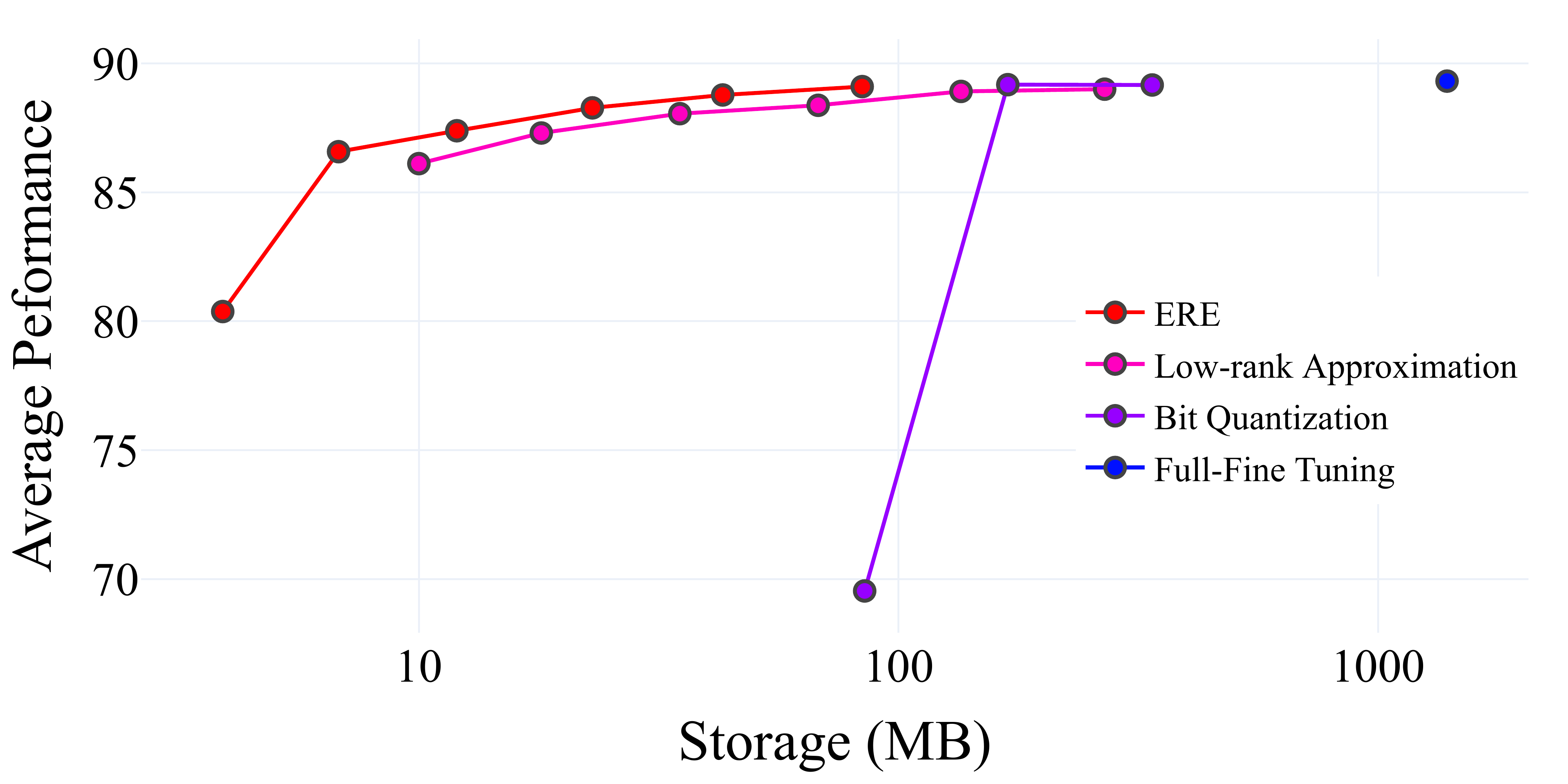}
    \caption{Average performance of RoBERTa-Large models with different weight residual compression methods on GLUE tasks vs storage usage. The storage requirement for the full weights is 1.36GB and the most competitive ERE, positioned at the rightmost of the red line, exhibits only a 0.2\% degradation in accuracy with 84MB. See \S\ref{nlu_experiment} for details.}
    \label{fig:teaser}
\end{figure}

Based on our findings, we propose Efficient Residual Encoding (ERE), which involves fully fine-tuning the model and efficiently storing weight residuals in a low-rank format. We also employ rank-wise vector quantization, using a low-rank approximation based on the weight-shiftedness of individual layers, measured through their approximated spectral distribution. The result of our method is demonstrated in Figure \ref{fig:teaser}.

We conducted experiments on multiple tasks, including Natural Language Understanding (NLU), Language Modeling (LM), and image generation, to evaluate the effectiveness of our proposed method.
In the NLU task, we demonstrated that the weight residuals can be compressed by up to 6.0\% of full fine-tuned weights while achieving comparable performance. Specifically, we achieved an average accuracy of 89.2 on the General Language Understanding Evaluation (GLUE) benchmark, which is very close to the performance of the original weights, 89.4.
We also demonstrate that ERE can successfully preserve the original generative quality with significant efficient storage in image generation and LM tasks.
We summarize our contributions below.

\begin{itemize}
    \item We observe that the differences in fine-tuned models, especially those with small parameter changes, can be approximated with significantly less memory than the base model.
    \item We further observe that different layers require different budgets of approximation. To this end, we propose ERE, a novel approach that enables the discriminatory allocation of resources among layers to obtain optimal storage.
    \item  As our method is model agnostic, we demonstrate the effectiveness of ERE across various tasks and modalities. Our method achieves efficient storage while maintaining performance.

\vspace{-3mm}
\end{itemize}

\section{Related Works}

\subsection{Transfer Learning}

Transfer learning involves fine-tuning a pre-trained model on a specific downstream task using a smaller labeled dataset. 
This approach capitalizes on the knowledge acquired during pre-training, enabling the model to quickly adapt to the new task with reduced data and computation requirements \cite{sharif2014cnn, kolesnikov2020big}. 
Transfer learning has proven to be highly effective in enhancing the performance of language models across various NLU tasks \cite{howard2018universal,bert,bart}.
Furthermore, there have been significant efforts to customize generative models, such as GANs \cite{ftgan1, ftgan2} and diffusion-based models \cite{ruiz2023dreambooth, textual_inversion}.

\subsection{Parameter Efficient Fine-Tuning}

Parameter Efficient Fine-Tuning (PEFT) methods have emerged as an alternative to achieve efficient transfer learning, as fully fine-tuning pre-trained model weights can be computationally expensive. 
LoRA is a prominent PEFT method that utilizes a bottleneck layer similar to adapter. 
It requires a predefined rank $r$ with $\alpha$ prior to train, which determines the reparameterized update scale. The fixed rank $r$ is uniformly applied to all layers, assuming equal budget requirements for each layer. 
Recent works like AdaLoRA \cite{zhangadaptive} and DyLoRA\cite{li2020dylora} have addressed this limitation by dynamically allocating the budget based on the importance of individual layers. While they are promising, they introduce even harder implementation difficulties. 

In contrast, our proposed method is not integrated into the training procedure itself. 
Therefore, once the model is trained, we have the flexibility to choose how the weight residuals are compressed based on the ERE factors. 

\section{Analysis On Weight Residuals}

\subsection{Effective Rank Dynamics of $\Delta \theta$}
\label{effrank}

\begin{figure}[ht!]
  \centering
    \includegraphics[width=1.0\textwidth]{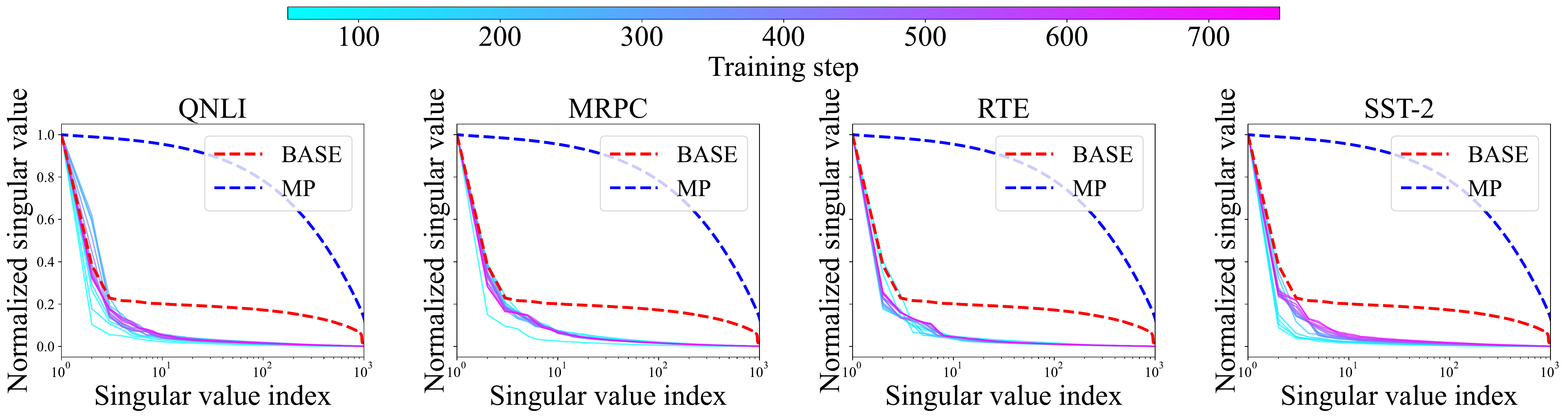}
    \caption{
    Normalized singular value magnitudes of weight residuals over time. 
    The red dotted line represents the spectral values of the base model, while the blue dotted line represents the expected spectral value quantile from the Marchenko-Pastur (M.P.) Distribution (λ = 0.25). 
    The M.P. Distribution serves as a suitable proxy for the theoretical singular-value distribution with random matrix initialization. }
    \label{fig:eigdist_prog}
\end{figure}

Let us denote the pre-trained model weight as $\theta$, and fine-tuned model as $\theta'$.
We conduct a detailed analysis of the training dynamics of the residual weights, denoted as $\Delta \theta = \theta ' - \theta $ during fine-tuning of the RoBERTa-Large model on various NLU tasks. We investigate the effective rank of the weight residuals and compare them to randomly initialized networks. Consistent with previous literatures \cite{huhlow, heavytailed}, we find that large overparameterized models, including pre-trained RoBERTa models, exhibit a low-rank structure as training progresses.

Interestingly, we observe that the weight residuals possess significantly lower effective ranks.
This is demonstrated in Figure \ref{fig:eigdist_prog}, where we plot the magnitude of the singular values normalized by the maximum singular value over time. The eigenvalue distribution rapidly converges to a low-rank regime, indicating a much smaller effective rank compared to the base parameters.

\begin{figure}[ht!]
  \centering
    \includegraphics[width=1.0\textwidth]{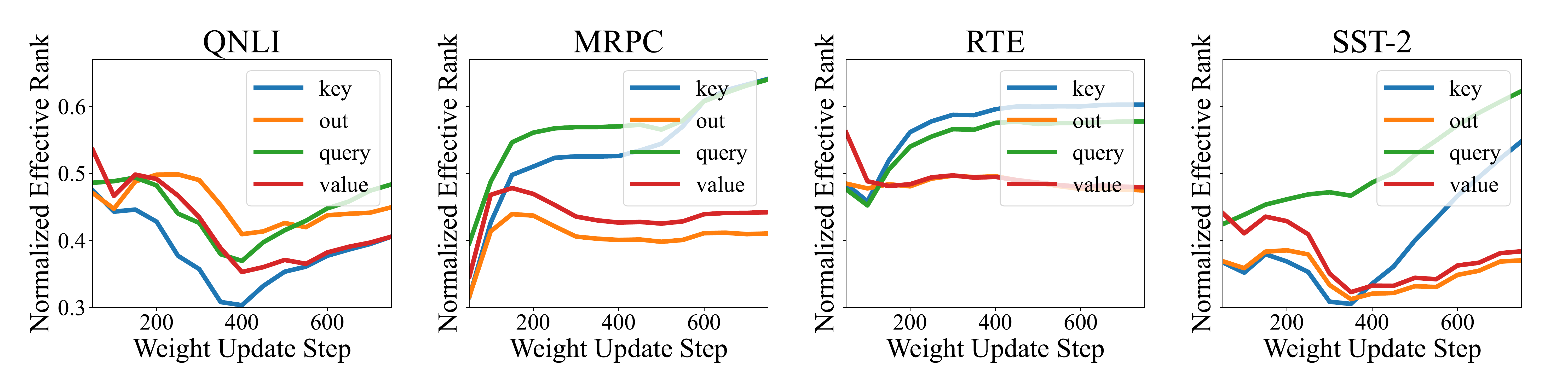}
    \caption{
    The changes of the effective rank of the query, key, value and output projection matrices from the last attention layer of the RoBERTa-Large model based on the weight update process. 
    The normalized effective ranks, $\text{erank}(\Delta \theta) / \text{erank}(\theta)$ of each layer are visualized per task.
    }
    \label{fig:noramlizederanktrend}
\vspace{-3.5mm}
\end{figure}

Furthermore, we discover that within the same layers, the effective ranks of the weight residuals are unpredictable and dynamic, as illustrated in Figure \ref{fig:noramlizederanktrend}. This finding helps explain the need for extensive hyper-parameter search and the inconsistent performance of PEFT across tasks, as demonstrated in previous works \cite{peftvsft1, peftvsft2, lora_oninstructgpt, zhangadaptive}.

Although similar behavior could have been hypothesized based on previous studies ~\cite{measureintrin, karimi2021compacter,lora, gradapprox1, kiani2022projunn, gooneratne2020low,aghajanyan2021intrinsic, huhlow, heavytailed}, to our best knowledge, our study is the first to systematically observe this behavior.

\subsection{Robustness of $\Delta \theta$}

\label{perturbable}

\begin{figure*}[t]
  \centering
  \begin{subfigure}[b]{0.49\textwidth}
    \includegraphics[width=\textwidth]{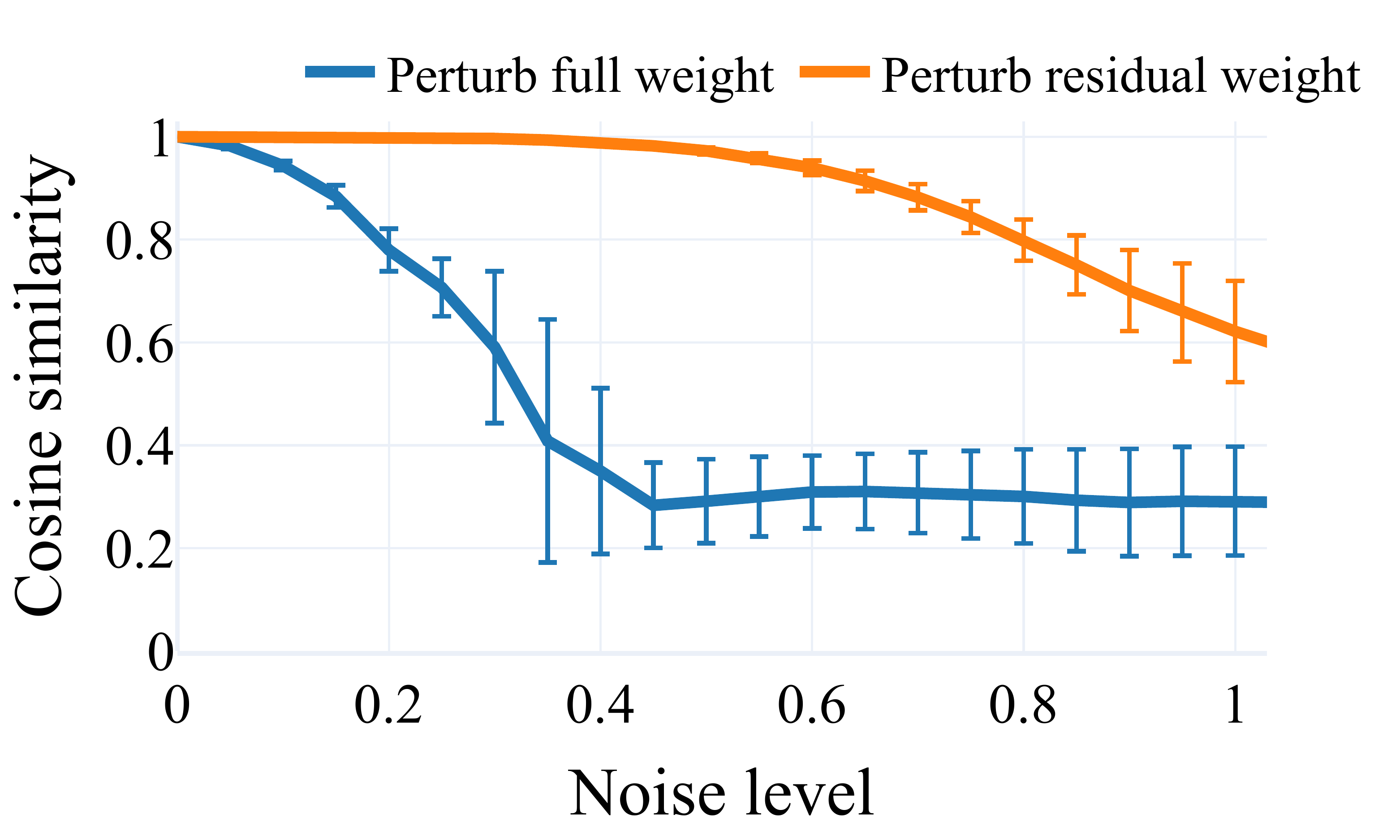}
    \caption{Log-normal perturbation}
    \label{fig:weight_perturb}
  \end{subfigure}
  \hfill
  \begin{subfigure}[b]{0.49\textwidth}
\includegraphics[width=\textwidth]{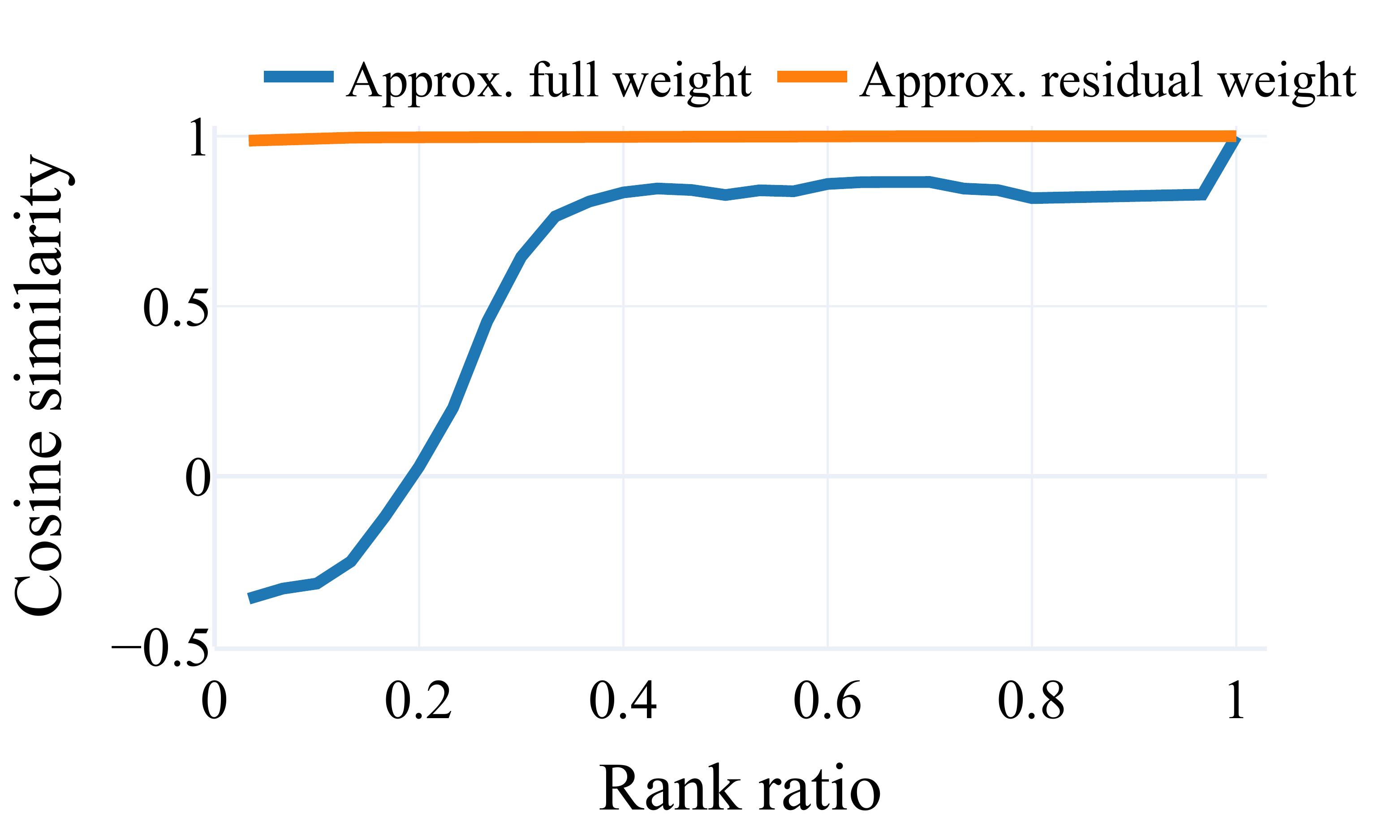}
    \caption{Low-rank approximation}
    \label{fig:weight_rank}
  \end{subfigure}
  \label{fig:comparison}
\caption{
Cosine similarity between feature outputs of perturbed weights $\widetilde{\theta'}$ and fine-tuned weights $\theta'$. The perturbation is applied to full fine-tuned weight ($\theta + \Delta \theta$) itself and weight residual ($\Delta \theta$) to compare the robustness to the perturbation. The perturbation is applied in two ways: (a) adding random noise, and (b) applying a low-rank approximation.
}

\vspace{-4mm}
\end{figure*}

We observed that fine-tuned weight residuals are significantly less sensitive to random noise compared to pre-trained weights. By introducing log-normal perturbations to each parameter, we evaluate the impact of noise on the model's output representation. We perturb either weight or weight residuals using \(\widetilde{\theta'} = (\theta + \Delta \theta)\exp(Z)\) or \(\widetilde{\theta'} = \theta + \exp(Z)\Delta \theta\), where \(Z \sim N(0, \sigma^2)\) and measure the change in representation using angular distance. From \autoref{fig:weight_perturb}, one can see that noise perturbation on the residual does little to no actual representation deviation to certain degree.

Building upon the previous findings regarding the low rankedness of weight residuals, we conclude that weight residuals are significantly redundant to their minimal impact. To illustrate this, we perform low-rank approximation on both the full weights and weight residuals, comparing the resulting differences in output representation using cosine similarity. \autoref{fig:weight_rank} highlights the stark contrast between the base parameters, which exhibit a substantial deviation when reducing 20\% of their rank, and the weight residuals, which do not introduce such drastic differences.

\vspace{-4.5mm}

\section{ERE: Efficient Residual Encoding}
Based on our observations, we introduce Efficient Residual Encoding (ERE) as a method for compressing weight residuals instead of full fine-tuned weights. By leveraging low-rank approximation techniques, guided by the requirements of individual layers (as observed in \S\ref{effrank}), we achieve even more efficient storage of the weight residuals.

\subsection{Low-rank Approximation and Budget Allocation}
\label{sec:4.1}
We first approximate $\Delta \theta$ using the product of three matrices, $U$, $D$, and $V$, where $U$ and $V$ are low-rank Stiefel matrices, and $D$ is a diagonal matrix. While this solution can be obtained through Singular Value Decomposition, some layers require significantly less rank to accurately approximate $\Delta \theta$. 
We can use the individual Empirical Spectral Distribution (ESD) to determine the appropriate amount of rank for $\Delta \theta$ after fine-tuning. Given a fixed parameter budget $M$, our objective is to distribute the rank in a way that minimizes the average Frobenius norm for the value. 
Let best possible rank $k$ approximation of weight residual $w = \Delta \theta$ be $w_k = \sum_{i = 1}^k \sigma_i u_i v_i^T$. Frobenius error by such approximation is
\begin{equation*}
  \lvert w - w_k \rvert_F = \lvert \sum_{i = k+1}^n 
  \sigma_i u_i v_i^T \rvert_F = \sqrt{ \sum_{i = k+1}^n \sigma_i^2}
\end{equation*}
where $w$ is a $m \times n$ matrix with $m \leq n$~\cite{eckart1936approximation}.
Our objective is to minimize the sum of every layer's low-rank approximation error. As each layer's rank $r$ comes with a parameter budget of $(n+m)r$, our objective can be expressed as follows:
\begin{align*}
     \min_{\boldsymbol{r}} &\sum_{i=1}^{N} \sum_{l=r_i+1}^{\min(n_i, m_i)} \sigma_l^2 && \text{(Total Squared Error)} \\
    \text{subject to}\quad & \sum_{i=1}^{N} r_i(n_i+m_i) \leq M && \text{(Budget)}
\end{align*}
Here, $n_i \times m_i$ are the shapes of the weight matrices of layer $i$, and $N$ is the number of all fine-tunable layers for the model. Note that this formulation is a special case of the well-known knapsack problem and, as such, is NP-Complete. Thus to find the approximate solution, we approximate $f_i(r) = \sum_{l = r + 1}^{min(n_i, m_i)} \sigma_l^2$ log-linearly based on layer $i$'s ESD, impose continuous relaxation on rank variables, and solve the relaxed problem in a Lagrangian formulation and binary search. See \S\ref{sec:appendix_a} for a detailed explanation of our algorithm.

\subsection{Further Optimization}
\label{further_opt}

While our general idea is described above, we can deploy a few modifications to further enhance performance. First, to address the potential overestimation of one effect of the rank over the other, we incorporate \textbf{prior rank}. This involves uniformizing the relaxed rank solution $\textbf{r} \in \mathbb{R}^N$ by taking $r_i ' = (1 - \alpha) r_i + \alpha r_{avg} $, where $r_{avg}$ is defined as $\sum_{i=1}^{N} r_{avg} (n_i+m_i) = M$. Note that this uniformization does not violate budget constraints. We have empirically determined $\alpha = 0.5$ to work well. This heuristic is necessary as our formulation, aimed at reducing the aggregate of Frobenius norm, is not perfectly aligned with the ultimate objective of reconstructing the final feature representation. A more sophisticated objective remains a potential future direction of research. 

Second, we observed that aggressive quantization can be applied without significant performance loss, consistent with findings in \cite{dettmers2022llm, yao2022zeroquant}. Following their approach, we utilized round-to-nearest quantization for $U$ and $V$, while reserving half-precision for $D$. However, we encountered a potential issue when dequantizing $U$ and $V$, as their precision might be significantly lower than that of $D$, violating our assumption that $U$ and $V$ are Stiefel elements. To address this, we conveniently project the dequantized $U$ and $V$ back onto the nearest Stiefel manifold. Further details and ablation results can be found in \S\ref{sec:appendix_b}.

\section{Experiments}
\label{sec:exp}

\begin{table}[ht!]
  \centering
  \footnotesize
  \addtolength{\tabcolsep}{-4pt}
  \begin{adjustbox}{width=0.85\textwidth}
  \begin{tabular}{l|r|ccccccccc}
  \hline
  \toprule
    Method   & Storage & MNLI & SST-2 & MRPC & CoLA & QNLI & QQP & RTE & STS-B & Avg. \\
  \midrule

  FT (\cite{liu2019roberta}) & 1.36GB & 90.2 & 96.4 & 90.9 & 68.0 & 94.7 & 92.2 & 86.6 & 92.4 & 88.9 \\
  \hb FT $^\ast$ & 1.36GB & 90.1 & \textbf{96.3} & \textbf{92.4} & 67.8 & \textbf{94.9} & \textbf{92.2} & 88.8 & \textbf{92.0 } & \textbf{89.4} \\
  \midrule
  BITQ (b=2) & 85MB & 37.9 & 94.0 & 84.1 & 51.9 & 92.1 & 36.8 & 83.0 & 88.0 & 71.0 \\
  BITQ (b=4) & 169MB & 90.1 & 96.2 & 92.4 & 67.8 & 94.9 & 92.2 & 88.8 & 92.0 & 89.3 \\

  \midrule
  
  LRA (r=8) & 20MB & 83.7 & 95.4 & 90.4 & 63.4 & 94.0 & 86.0 & 84.8 & 91.6 & 86.2 \\
  LRA (r=16) & 36MB & 86.1 & 95.6 & 91.4 & 65.4 & 94.6 & 86.9 & 87.4 & 91.8 & 87.4 \\
  LRA (r=32) & 70MB & 88.6 & 96.0 & 91.4 & 66.1 & 94.8 & 88.2 & 88.4 & 91.9 & 88.2 \\
  LRA (r=64) & 136MB & 89.6 & 96.0 & 91.9 & 65.8 & 94.7 & 89.8 & 88.1 & 92.0 & 88.5 \\
  LRA (r=128) & 270MB & 90.0 & 96.0 & 91.9 & 67.5 & 94.8 & 91.3 & 88.8 & 92.0 & 89.0 \\
  LRA (r=256) & 538MB & 90.1 & 96.1 & 92.2 & 67.7 & 94.7 & 92.0 & 88.1 & 92.0 & 89.1 \\
  % \midrule
  % ERE (r=4, b=2) & 2.3MB & 31.8 & 93.3 & 88.0 & 61.1 & 91.7 & & 67.5 & 88.6 & \\
  % ERE (r=8, b=2) & 3.0MB & 39.5 & 95.2 & 89.0 & 61.3 & 93.0 & 76.3 & 86.6 & 90.2 & 78.9 \\
  % ERE (r=16, b=2) & 4.4MB & 81.4 & 95.3 & 89.2 & 63.1 & 93.8 & 84.6 & 87.0 & 90.7 & 85.6 \\
  % ERE (r=32, b=2) & 7.3MB & 85.3 & 95.3 & 89.7 & 65.3 & 94.2 & 86.4 & 88.8 & 90.9 & 87.0 \\
  % ERE (r=64, b=2) & 13MB & 88.0 & 95.4 & 90.2 & 65.3 & 94.4 & 88.2 & 88.1 & 91.0 & 87.6 \\
  % ERE (r=128, b=2) & 25MB & 89.2 & 95.6 & 90.2 & 66.5 & 94.6 & 89.9 & 88.4 & 91.1 & 88.2 \\
  % \hb ERE (r=256, b=2) & 48MB & 89.5 & 95.8 & 90.2 & 67.8 & 94.5 & 91.2 & 88.4 & 91.1 & 88.6 \\
  \midrule
  ERE (r=8, b=4) & 3.9MB & 47.7 & 95.2 & 90.4 & 63.0 & 92.8 & 75.9 & 86.6 & 90.9 & 80.3 \\
  ERE (r=16, b=4) & 6.8MB & 83.9 & 95.4 & 90.7 & 65.0 & 94.1 & 85.2 & 87.4 & 91.4 & 86.6 \\
  ERE (r=32, b=4) & 12MB & 86.9 & 95.6 & 91.4 & 64.6 & 94.6 & 87.1 & 88.1 & 91.5 & 87.5 \\
  ERE (r=64, b=4) & 23MB & 89.1 & 95.6 & 91.9 & 66.5 & 94.7 & 88.7 & \textbf{89.2} & 91.7 & 88.4 \\
  ERE (r=128, b=4) & 43MB& 90.0 & 95.9 & 91.9 & 67.2 & 94.7 & 90.6 & 89.2 & 91.7 & 88.9 \\
  \hb ERE (r=256, b=4) & 84MB & \textbf{90.2} & 96.1 & 92.2 & \textbf{68.2} & 94.7 & 91.7 & 88.8 & 91.8 & 89.2 \\
  \bottomrule
  \end{tabular}
  \end{adjustbox}
  \caption{
  The comparison between $\text{RoBERTa-Large}$ with full Fine-Tuning (FT) method and ERE on GLUE.
  Note that, the baseline we implement($^{\ast}$) is slightly better than the reported results in the paper.
  We measure Mattew's correlation for CoLA Pearson correlation for STS-B, and overall accuracy for other tasks.
  Additionally, we provide the results when applying BIT Quantization (BITQ), and Low-Rank Approximation (LRA) independently. 
  $r$ and $b$ denote rank and quantization level each.
  }
  \label{tab:nlu_table}
\vspace{-3.5mm}
\end{table}

\subsection{ERE on Natural Language Understanding}
\label{nlu_experiment}
To assess the effectiveness of our method, ERE, we first conducted evaluations on the standard NLU benchmark, GLUE \citep{wang-etal-2018-glue}.
We fine-tune the provided pre-trained RoBERTa-Large~\cite{liu2019roberta} model on each individual task, employing a single-model and single-task fine-tuning setup. 
We train and evaluate each model with Fairseq\footnote{\url{https://github.com/facebookresearch/fairseq}} \citep{ott2019fairseq} framework.
We utilize the official training configurations to reproduce the performance of the full fine-tuning model and apply ERE on the weight residuals between pre-trained and fine-tuned models. 
We conduct evaluations using various rank settings for ERE and report both the overall accuracy and the corresponding storage requirements.
Note that, we save model weights with full precision except newly added classifiers because they have no counterparts to compress residual.
We also report the results when BIT Quantization (BITQ) and Low-Rank Approximation (LRA) techniques are applied independently. 
For all the experiments applied ERE, $r$ indicates prior rank, and $b$ means bit quantization level.
Note that, the bit quantization technique used in BITQ differs from ERE. In ERE, only the $U$ and $V$ components of the low-rank factorized residuals are quantized as mentioned \S\ref{further_opt}, while BITQ quantizes all elements of the weight residuals.

\autoref{tab:nlu_table} illustrates that ERE (r=256, b=4) results in a significantly reduced model, requiring only 6.0\% of the original storage space, with minimal impact on performance degradation.
The results indicate that residual compression can be effectively achieved even when employing only BITQ or LRA techniques. However, it is important to note that these methods require larger storage compared to ERE.
Nevertheless, despite the advantages of our method in efficient storage of model weights, there exists a potential for failure in accurately reconstructing the original model's representations when compressing residuals with extremely low rank and bit, particularly in tasks such as MNLI and QQP.
We hypothesize that this is because these tasks take much larger weight updates compared to other tasks, where MNLI and QQP takes about 37 times larger training wall clock time on a single V100 GPU machine, while maintaining the same batch size. 
Another possibility could be that fine-tuning \text{RoBERTa} with NLU tasks requires a newly initialized classifier unlike other tasks such as text-to-image generation and language modeling.
Furthermore, it is worth noting that the application of ERE yields improved performance in certain tasks (MNLI, CoLA, and RTE). This improvement can be attributed to the elimination of redundant or performance-degrading factors through the encoding of weight residuals with ERE.

\begin{figure}[ht!]
  \centering
    \includegraphics[width=0.8\textwidth]{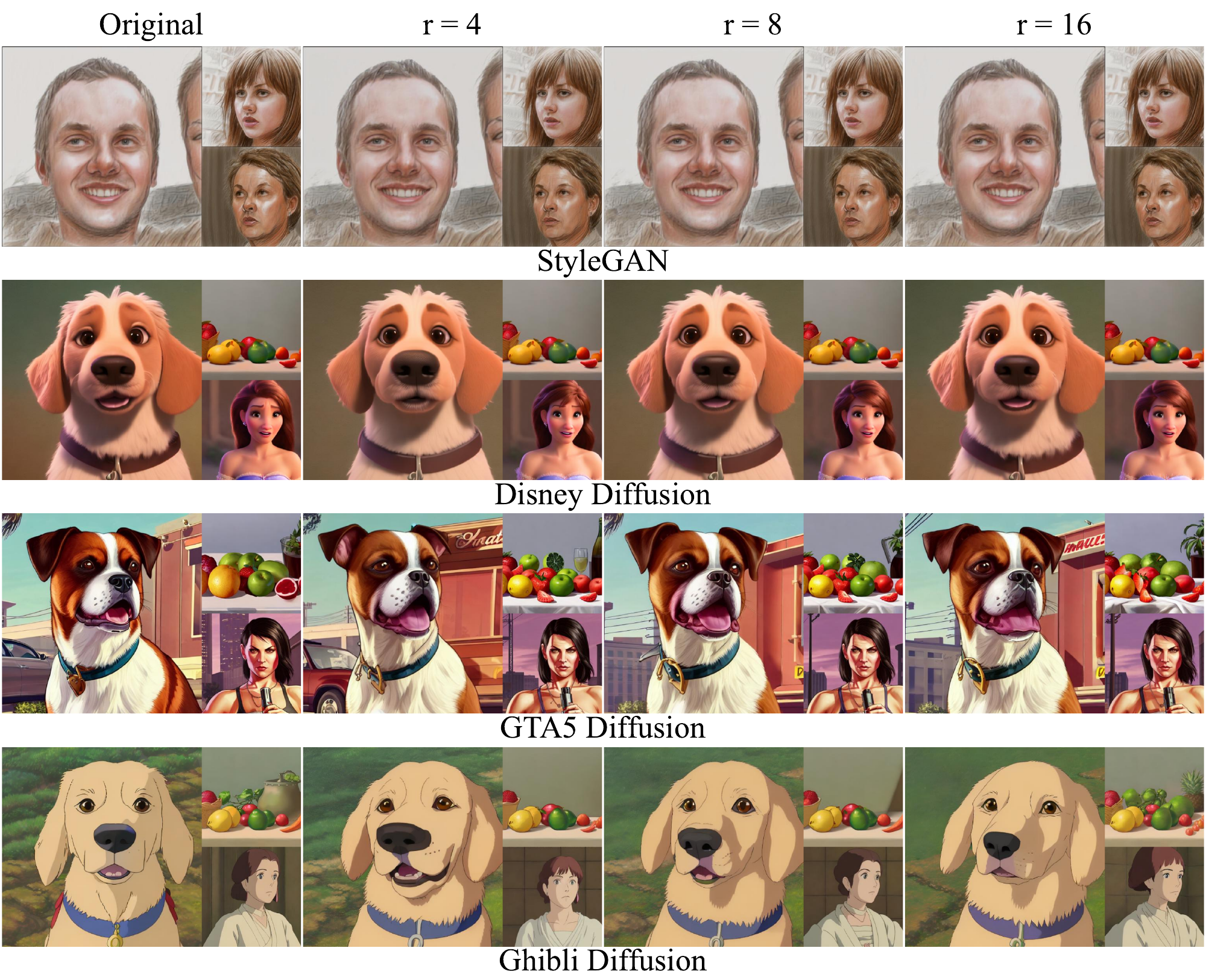}
    \caption{
    The comparison between generated samples of fine-tuned StyleGAN with NADA method and Latent Diffusion Model and their ERE reconstructed counterparts. 
    Diffusion-based models generate images based on three prompts: "A portrait of a dog in $<Style>$", "A portrait of a woman in $<Style>$" and "A still life painting of fruit in $<Style>$", where the $<Style>$ token is replaced with each of the styles: $disney$, $gtav$, and $ghibli$.
    The ERE-applied models are compressed using 4-bit quantization and has prior ranks of 4, 8, and 16, respectively.
    }
    \label{fig:image_generation}
\end{figure}

\begin{table}[ht!]
\centering
\begin{adjustbox}{width=0.9\textwidth}
\begin{tabular}{c|ccc|c}
\hline
Model Type & ERE (r=4, b=4) & ERE (r=8, b=4) & ERE (r=16, b=4) & FT \\
\hline
StyleGAN & 0.305 (0.46MB) & 0.291 (0.67MB) & 0.281 (1.08MB) & - (364MB) \\
GTA5 Diffusion & 0.365 (3.7MB) & 0.327 (6.1MB) & 0.305 (11MB) & - (3.44GB) \\
Disney Diffusion & 0.221 (3.7MB) & 0.193 (6.1MB) & 0.159 (11MB) & - (3.44GB) \\
Ghibli Diffusion & 0.345 (3.7MB) & 0.326 (6.1MB) & 0.299 (11MB) & - (3.44GB) \\
\hline
\end{tabular}
\end{adjustbox}
\caption{
LPIPS distance between fully fine-tuned models and their ERE applied counterparts. 
Lower is better. 
Note that, for the diffusion-based models, only the UNet module is considered for storage computation.
}
\label{tab:image_generation}
\vspace{-5mm}
\end{table}

\subsection{ERE on Image Generation}
This section presents our experiments on the image generation task using ERE. We utilize publicly available pre-trained and fully fine-tuned Latent Diffusion Models (LDMs)~\cite{LDM} and StyleGANs~\cite{karras2019style}.

For the diffusion-based models, we apply ERE on publically available fine-tuned models such as Disney diffusion\footnote{\url{https://huggingface.co/nitrosocke/mo-di-diffusion}}, GTA5 diffusion\footnote{\url{https://huggingface.co/ItsJayQz/GTA5_Artwork_Diffusion}}, Ghibli diffusion\footnote{\url{https://huggingface.co/nitrosocke/Ghibli-Diffusion}}. 
For the diffusion-based models, we fixed seed with 0 and 50 inference steps and a 7.5 guidance scale for sampling.
We randomly sample 40 images from a fixed set of prompts, and measure Learned Perceptual Image Patch Similarity (LPIPS) distance \cite{lpips} on generated images to measure the accurate reconstruction of the fine-tuned models.
Similarly for StyleGAN models, we perform CLIP-driven adaptation via NADA method ~\cite{gal2022stylegan} and compress the adapted models with ERE. 
We also sample 40 images with fixed seed 0 and measure their LPIPS distance. The results, as shown in ~\autoref{fig:image_generation} and ~\autoref{tab:image_generation}, demonstrate that ERE (r=4, b=4) can successfully preserve most of the semantic information. 
Furthermore, increasing the number of ranks leads to improved perceptual quality.

\subsection{ERE on Language Modeling}

\begin{table}[ht!]
  \centering
  \footnotesize
  \addtolength{\tabcolsep}{-4pt}
  \begin{adjustbox}{width=\textwidth}
  \begin{tabular}{l|c|c|c|c|c|c|c|c|c|c|c}
  \hline
  \toprule
    Method   & Storage & OpenBook & ARC & Wino & Hella & ARC & PIQA & BoolQ & OpenAI  & Avg. & Avg. \\
             &         & QA       & Easy & Grande & Swag & Challenge & & & Lambada & (w/o lambada) & (w/ lambada) \\
  \midrule
  \text{Pythia-2.8B}  & 10.7GB & 34.80 & 58.59 & 56.96 & 59.12 & 32.34 & 73.40 & 63.82 & - & 52.34 & - \\
  \text{Dolly-3B}   & 10.7GB & 38.40 & 61.15 & 58.96 & 65.08 & 37.03 & 74.27 & 57.55 & - & 54.49 & - \\
  \midrule
  \text{Pythia-2.8B}$^{\ast}$ & 10.7GB & 35.60 & 58.96 & 59.75 & 59.30 & 33.02 & 73.61 & 64.59 & 64.70 & 52.95 & 54.30 \\
  \hb \text{Dolly-3B}$^{\ast}$ & 10.7GB & 38.80 & 61.62 & 59.19 & 64.99 & 37.03 & 74.16 & 57.55 & 62.86 & 54.64 & 55.60 \\
  \midrule
  ERE (r=4, b=4) & 8.0MB & 38.00 & \textbf{63.76} & 59.43 & 64.54 & \textbf{35.75} & 75.68 & \textbf{61.62} & \textbf{65.98} & 56.97 & 58.10 \\
  \hb ERE (r=8, b=4) & 12.0MB & 38.00 & 63.72 & \textbf{59.75} & 65.01 & \textbf{35.75} & \textbf{75.90} & 60.92 & 65.79 & \textbf{57.01} & \textbf{58.11} \\
  ERE (r=16, b=4) & 20.0MB & 38.40 & 63.47 & 59.59 & 65.10 & 35.67 & 75.57 & 61.22 & 65.67 & 57.00 & 58.09 \\
  ERE (r=32, b=4) & 35.0MB & \textbf{39.00} & 63.01 & 59.27 & \textbf{65.36} & 35.58 & 75.46 & 59.85 & 65.22 & 56.79 & 57.84 \\
  \midrule
  \midrule
  \text{Pythia-6.9B}  & 26.3GB & 36.80 & 60.48 & 60.85 & 63.15 & 34.39 & 76.12 & 62.63 & - & 54.36 & - \\
  \text{Dolly-7B} & 26.3GB & 39.20 & 63.39 & 60.77 & 68.65 & 40.70 & 75.08 & 64.40 & - & 57.35 & - \\
  \midrule
  \text{Pythia-6.9B}  $^{\ast}$ & 26.3GB & 37.20 & 61.24 & 60.69 & 63.85 & 35.32 & 76.44 & 63.30 & \textbf{67.09} & 56.86 & 58.14 \\
  \hb \text{Dolly-7B} $^{\ast}$ & 26.3GB & \textbf{39.80} & 63.05 & 61.64 & 68.56 & 39.59 & 74.43 & 64.92 & 62.74 & 58.86 & 59.34 \\
  \midrule
  \hb ERE (r=4, b=4) & 13.0MB & \textbf{39.80} & \textbf{65.91} & \textbf{62.98} & 69.12 & 38.82 & \textbf{77.04} & 64.92 & 67.01 & 59.80 & \textbf{60.70} \\
  ERE (r=8, b=4) & 19.0MB & 39.20 & 65.78 & 61.88 & \textbf{69.48} & 39.59 & 76.82 & \textbf{65.72} & 66.39 & 59.78 & 60.60 \\
  ERE (r=16, b=4) & 31.0MB & 39.40 & 65.57 & 62.27 & 69.42 & \textbf{40.36} & \textbf{77.04} & 65.02 & 66.25 & \textbf{59.87} & 60.67 \\
  ERE (r=32, b=4) & 55.0MB & 39.40 & 65.19 & 62.35 & 69.20 & 39.93 & 76.28 & 64.80 & 65.59 & 59.59 & 60.34 \\
  \bottomrule

  \end{tabular}
  \end{adjustbox}
  \caption{
  The comparison between pre-trained fine-tuned, and ERE applied models on zero-shot evaluation benchmarks. Performance is measured using normalized accuracy for OpenBookQA, ARC-Easy, HellaSwag, ARC-Challenge, and PIQA, and accuracy for other tasks, with the average accuracy calculated across all tasks. ($^{\ast}$) denotes our implementation using the provided model checkpoints. $r$ and $b$ represent the rank and quantization level, respectively. 
  In our storage computation, we exclude the word embedding layers because the shapes of the embedding layers in the fine-tuned model differ from the pre-trained model.
  }
  \label{tab:llm_zero_shot}
\vspace{-3.5mm}
\end{table}

In this section, we present the evaluation of ERE on the Large Language Model (LLM), Dolly\footnote{\url{https://github.com/databrickslabs/dolly}}. Dolly is a fully fine-tuned model based on Pythia \citep{biderman2023pythia}, utilizing a dataset consisting of approximately 15k instructions. 
We evaluate two models with 3 billion (3B) and 7 billion (7B) parameters with 8 zero-shot evaluation tasks using open-source LM evaluation framework\citep{leo_gao_2022_7413426}.

\autoref{tab:llm_zero_shot} presents the results of applying ERE to the fully fine-tuned LLM model. It shows that the model with ERE outperforms the original model in zero-shot evaluation tasks. However, there is a noticeable trend of performance decline beyond a certain rank. Specifically, for the 7B model, the performance begins to decline after rank 4, while for the 3B model, it starts to decline after rank 8.
We hypothesize this is because the original purpose of the dolly is to follow instruction prompts, rather than achieving zero-shot performances which is the inherent capability of the pre-trained baseline. 
Thus, we interpret this phenomenon as ERE enabling the restoration of the model's generalizability while preserving its ability to understand instruction prompts. 
Supporting this interpretation, Figure \ref{fig:llm_inference} demonstrates that the model with ERE retains the capacity to understand instruction prompts, whereas the pre-trained baseline does not. 

\begin{figure}[ht!]
{
\tt 
% \centering
\footnotesize
\begin{center}
\begin{adjustbox}{width=0.9\textwidth}
\begin{tabularx}{\linewidth}{r X}
\toprule
Instruction $\to$ & Explain to me Convolutional Neural Networks (CNNs) \\
 & Response: \\
\midrule
Pre-trained $\to$ & Response:

Response:

Response:

\color{red}{(Skip)} \\

\midrule
FT (26.3GB) $\to$ & CNNs are a type of neural network that are popular for use in image processing applications such as computer vision. CNNs are comprised of convolutional and non-linear layers that are designed to process spatial information within an image. \color{red}{(Skip)} \\ 
% & End \\
\midrule
ERE (13MB) $\to$ & CNNs are a type of neural network that are popular in computer vision, as they can perform spatial and temporal processing of an input image that would be too difficult for other types of neural networks. \color{red}{(Skip)} \\
% & End \\
\bottomrule
\end{tabularx}

\end{adjustbox}
\end{center}
}
\caption{
Example of generated samples from instruction prompt. We fix the random seed and use the default configuration (same top-p and top-k probability) for sampling. Note that pre-trained model return 51 repetitive sentences "Response: Response: ...". because it has no capability to understand the instruction prompt, while ERE can understand the instruction prompt and response like the fully fine-tuned model. The responses over a certain length are skipped for brevity.
}
\label{fig:llm_inference}
\vspace{-3.5mm}
\end{figure}

\section{Discussions and Limitations}

\subsection{Why do we need ERE when we have PEFT?}

In this section, we emphasize that while PEFT has shown impressive performance, there are situations where reverting to full fine-tuning is still necessary, making ERE a valuable alternative.

Firstly, integrating other PEFT methods that disrupt the computational graph at inference time can be challenging and costly. Many PEFT methods modify the flow of computing variables and are specifically designed for a single model architecture. For instance, integrating prefix tuning \cite{li2021prefix} with Paint-with-words \cite{balaji2022ediffi} or Adapters \cite{adapter} with ControlNet \cite{controlnet} poses significant difficulties due to their architectural incompatibilities. In contrast, full fine-tuning does not introduce any additional complexity in terms of compatibility.

Secondly, by solely relying on PEFT, users may overlook extensive research conducted under the assumption of full-weight fine-tuning. This research includes dedicated fields, such as prior-focused continual learning \cite{ewc, vcl, synint, aljundi2018memory}, and weight-averaging methods \cite{fedavg, robustft, modelsoup, matena2022merging}.

\subsection{The Comparison between ERE and LoRA}

We undertake a comparative analysis between ERE and LoRA~\cite{lora}, aiming to delineate the strengths and weaknesses of both methods.
Both methods share a common characteristic in leveraging the low-rank nature of weight residuals ($\Delta \theta$).
For a fair comparison, we implement LoRA in Fairseq using official implementation\footnote{\url{https://github.com/microsoft/LoRA}}.
We first fully fine-tuned the model for each GLUE tasks and compute the total train wall clock. 
Then we train LoRA injected RoBERTa-Large model within the same training time budget. We follow most of the hyperparameters in the LoRA paper, and initialize the classifier parameters with the same seed and use the same GPU (V100), framework (Fairseq), and PyTorch version (2.0).

In \autoref{tab:ere_vs_lora}, we observe that the overall accuracy of LoRA is comparable to that of the fully fine-tuned model. 
However, both ERE and the fully fine-tuned model achieve better performances. 
On the other hand, LoRA requires smaller storage compared to ERE for achieving competitive performance.
However, it is important to note that a direct comparison with ERE and LoRA may be challenging because the full fine-tuning method updates more layers including the word embedding layer and bias.
The fully fine-tuned model weight can be stored efficiently with ERE, requiring 6.0\% of the original storage while preserving similar performance and 1.7\% for the same performance compared to LoRA.

As a result, depending on the specific task and requirements, one might choose LoRA due to its competitive performance and smaller storage requirements. However, in certain scenarios where maximizing performance is crucial, the full fine-tuning approach and ERE could be preferred. Furthermore, it is worth noting that ERE can be applied to weights obtained through LoRA, providing additional flexibility and options for achieving optimal trade-offs between performance and storage efficiency. This highlights the versatility and potential benefits of ERE in combination with PEFT methods.

\begin{table}[ht!]
  \centering
  \footnotesize
  \addtolength{\tabcolsep}{-4pt}
  \begin{adjustbox}{width=0.85\textwidth}
  \begin{tabular}{l|r|cccccccccc}
  \hline
  \toprule
    Method   & Storage & MNLI & SST-2 & MRPC & CoLA & QNLI & QQP & RTE & STS-B & Avg.  \\
  \midrule

  FT & 1.36GB & 90.1 & 96.3 & \textbf{92.4} & 67.8 & \textbf{94.9} & \textbf{92.2} & 88.8 & 92.0 & \textbf{89.4} \\
  \hb ERE (r=256, b=4) & 84MB & \textbf{90.2} & 96.1 & 92.2 & \textbf{68.2} & 94.7 & 91.7 & 88.8 & 91.8 & 89.2 \\
  ERE (r=64, b=4) & 23MB & 89.1 & 95.6 & 91.9 & 66.5 & 94.7 & 88.7 & \textbf{89.2} & 91.7 & 88.4 \\

  \midrule
  LoRA (r=8)~\cite{lora} & 3.0MB & 90.6 & 96.2 & 90.9 & 68.2 & 94.9 & 91.6 & 87.4 & 92.6 & 89.0 \\
  LoRA (r=8)$^{\ast}$ & 3.0MB & 90.3 & 96.3 & 91.7 & \textbf{68.6} & 94.7 & 90.9 & 88.4 & 91.8 & 89.1 \\
  \hb LoRA (r=8)$^{\dagger}$ & 3.0MB & \textbf{90.3} & 96.3 & 89.7 & 65.8 & 94.7 & 90.5 & 88.4 & 91.4 & 88.4 \\
  \bottomrule
  \end{tabular}
  \end{adjustbox}
  \caption{
  The comparison between ERE and LoRA. ($^{\ast}$) denotes our LoRA implementation without training time constraint, while ($^{\dagger}$) indicates LoRA with an equal training time budget constraint.
  The newly added linear layer for the downstream finetuning task is not considered for storage computation.
  }
  \label{tab:ere_vs_lora}
\vspace{-3.5mm}
\end{table}

\subsection{Limitations}

The limitations of our paper include the potential challenges of ERE in extreme compression scenarios, particularly when the model size is not large enough. This limitation becomes more prominent with a high number of updates or a high learning rate relative to the model size. Additionally, while we demonstrate the effectiveness of our method across various tasks and model architectures, we have not yet established its applicability in diverse fine-tuning configurations. We acknowledge the need for further investigation in these setups as a future research avenue.

\section{Conclusion}

This paper provides a thorough analysis of weights residual and introduces ERE as an effective approach for efficient storage without sacrificing performance. Our experiments demonstrate the efficacy of ERE across tasks like NLU, LM, and image generation. We believe ERE offers a compelling alternative to PEFT in terms of reducing footprint, due to their flexibility, simplicity, and competitive performance.

% \begin{ack}s
% Use unnumbered first level headings for the acknowledgments. All acknowledgments
% go at the end of the paper before the list of references. Moreover, you are required to declare
% funding (financial activities supporting the submitted work) and competing interests (related financial activities outside the submitted work).
% More information about this disclosure can be found at: \url{https://neurips.cc/Conferences/2023/PaperInformation/FundingDisclosure}.

% Do {\bf not} include this section in the anonymized submission, only in the final paper. You can use the \texttt{ack} environment provided in the style file to autmoatically hide this section in the anonymized submission.
% \end{ack}

\bibliographystyle{plain}
\bibliography{references}

\begin{thebibliography}{10}

\bibitem{aghajanyan2021intrinsic}
Armen Aghajanyan, Sonal Gupta, and Luke Zettlemoyer.
\newblock Intrinsic dimensionality explains the effectiveness of language model
  fine-tuning.
\newblock In {\em Proceedings of the 59th Annual Meeting of the Association for
  Computational Linguistics and the 11th International Joint Conference on
  Natural Language Processing (Volume 1: Long Papers)}, pages 7319--7328, 2021.

\bibitem{aljundi2018memory}
Rahaf Aljundi, Francesca Babiloni, Mohamed Elhoseiny, Marcus Rohrbach, and
  Tinne Tuytelaars.
\newblock Memory aware synapses: Learning what (not) to forget.
\newblock In {\em Proceedings of the European conference on computer vision
  (ECCV)}, pages 139--154, 2018.

\bibitem{baevski2020wav2vec}
Alexei Baevski, Yuhao Zhou, Abdelrahman Mohamed, and Michael Auli.
\newblock wav2vec 2.0: A framework for self-supervised learning of speech
  representations.
\newblock {\em Advances in neural information processing systems},
  33:12449--12460, 2020.

\bibitem{balaji2022ediffi}
Yogesh Balaji, Seungjun Nah, Xun Huang, Arash Vahdat, Jiaming Song, Karsten
  Kreis, Miika Aittala, Timo Aila, Samuli Laine, Bryan Catanzaro, et~al.
\newblock ediffi: Text-to-image diffusion models with an ensemble of expert
  denoisers.
\newblock {\em arXiv preprint arXiv:2211.01324}, 2022.

\bibitem{bitfit}
Elad Ben~Zaken, Yoav Goldberg, and Shauli Ravfogel.
\newblock {B}it{F}it: Simple parameter-efficient fine-tuning for
  transformer-based masked language-models.
\newblock In {\em Proceedings of the 60th Annual Meeting of the Association for
  Computational Linguistics (Volume 2: Short Papers)}, pages 1--9, Dublin,
  Ireland, May 2022. Association for Computational Linguistics.

\bibitem{biderman2023pythia}
Stella Biderman, Hailey Schoelkopf, Quentin Anthony, Herbie Bradley, Kyle
  O'Brien, Eric Hallahan, Mohammad~Aflah Khan, Shivanshu Purohit, USVSN~Sai
  Prashanth, Edward Raff, et~al.
\newblock Pythia: A suite for analyzing large language models across training
  and scaling.
\newblock {\em arXiv preprint arXiv:2304.01373}, 2023.

\bibitem{peftvsft1}
Guanzheng Chen, Fangyu Liu, Zaiqiao Meng, and Shangsong Liang.
\newblock Revisiting parameter-efficient tuning: Are we really there yet?
\newblock {\em arXiv preprint arXiv:2202.07962}, 2022.

\bibitem{dettmers2022llm}
Tim Dettmers, Mike Lewis, Younes Belkada, and Luke Zettlemoyer.
\newblock Llm. int8 (): 8-bit matrix multiplication for transformers at scale.
\newblock {\em Advances in Neural Information Processing Systems}, 2022.

\bibitem{bert}
Jacob Devlin, Ming{-}Wei Chang, Kenton Lee, and Kristina Toutanova.
\newblock {BERT:} pre-training of deep bidirectional transformers for language
  understanding.
\newblock In Jill Burstein, Christy Doran, and Thamar Solorio, editors, {\em
  Proceedings of the 2019 Conference of the North American Chapter of the
  Association for Computational Linguistics: Human Language Technologies,
  {NAACL-HLT} 2019, Minneapolis, MN, USA, June 2-7, 2019, Volume 1 (Long and
  Short Papers)}, pages 4171--4186. Association for Computational Linguistics,
  2019.

\bibitem{eckart1936approximation}
Carl Eckart and Gale Young.
\newblock The approximation of one matrix by another of lower rank.
\newblock {\em Psychometrika}, 1(3):211--218, 1936.

\bibitem{unitary}
Ky~Fan and Alan~J Hoffman.
\newblock Some metric inequalities in the space of matrices.
\newblock {\em Proceedings of the American Mathematical Society},
  6(1):111--116, 1955.

\bibitem{textual_inversion}
Rinon Gal, Yuval Alaluf, Yuval Atzmon, Or~Patashnik, Amit~H Bermano, Gal
  Chechik, and Daniel Cohen-Or.
\newblock An image is worth one word: Personalizing text-to-image generation
  using textual inversion.
\newblock {\em arXiv preprint arXiv:2208.01618}, 2022.

\bibitem{gal2022stylegan}
Rinon Gal, Or~Patashnik, Haggai Maron, Amit~H Bermano, Gal Chechik, and Daniel
  Cohen-Or.
\newblock Stylegan-nada: Clip-guided domain adaptation of image generators.
\newblock {\em ACM Transactions on Graphics (TOG)}, 41(4):1--13, 2022.

\bibitem{leo_gao_2022_7413426}
Leo Gao, Jonathan Tow, Stella Biderman, Charles Lovering, Jason Phang, Anish
  Thite, Fazz, Niklas Muennighoff, Thomas Wang, sdtblck, tttyuntian,
  researcher2, Zdeněk Kasner, Khalid Almubarak, Jeffrey Hsu, Pawan~Sasanka
  Ammanamanchi, Dirk Groeneveld, Eric Tang, Charles Foster, kkawamu1, xagi dev,
  uyhcire, Andy Zou, Ben Wang, Jordan Clive, igor0, Kevin Wang, Nicholas Kross,
  Fabrizio Milo, and silentv0x.
\newblock Eleutherai/lm-evaluation-harness: v0.3.0, December 2022.

\bibitem{gooneratne2020low}
Mary Gooneratne, Khe~Chai Sim, Petr Zadrazil, Andreas Kabel, Fran{\c{c}}oise
  Beaufays, and Giovanni Motta.
\newblock Low-rank gradient approximation for memory-efficient on-device
  training of deep neural network.
\newblock In {\em ICASSP 2020-2020 IEEE International Conference on Acoustics,
  Speech and Signal Processing (ICASSP)}, pages 3017--3021. IEEE, 2020.

\bibitem{adapter}
Neil Houlsby, Andrei Giurgiu, Stanislaw Jastrzebski, Bruna Morrone, Quentin
  de~Laroussilhe, Andrea Gesmundo, Mona Attariyan, and Sylvain Gelly.
\newblock Parameter-efficient transfer learning for {NLP}.
\newblock {\em CoRR}, abs/1902.00751, 2019.

\bibitem{howard2018universal}
Jeremy Howard and Sebastian Ruder.
\newblock Universal language model fine-tuning for text classification.
\newblock {\em arXiv preprint arXiv:1801.06146}, 2018.

\bibitem{lora}
Edward~J Hu, yelong shen, Phillip Wallis, Zeyuan Allen-Zhu, Yuanzhi Li, Shean
  Wang, Lu~Wang, and Weizhu Chen.
\newblock Lo{RA}: Low-rank adaptation of large language models.
\newblock In {\em International Conference on Learning Representations}, 2022.

\bibitem{huhlow}
Minyoung Huh, Hossein Mobahi, Richard Zhang, Brian Cheung, Pulkit Agrawal, and
  Phillip Isola.
\newblock The low-rank simplicity bias in deep networks.

\bibitem{karimi2021compacter}
Rabeeh Karimi~Mahabadi, James Henderson, and Sebastian Ruder.
\newblock Compacter: Efficient low-rank hypercomplex adapter layers.
\newblock {\em Advances in Neural Information Processing Systems},
  34:1022--1035, 2021.

\bibitem{karras2019style}
Tero Karras, Samuli Laine, and Timo Aila.
\newblock A style-based generator architecture for generative adversarial
  networks.
\newblock In {\em Proceedings of the IEEE/CVF conference on computer vision and
  pattern recognition}, pages 4401--4410, 2019.

\bibitem{kiani2022projunn}
Bobak Kiani, Randall Balestriero, Yann LeCun, and Seth Lloyd.
\newblock projunn: efficient method for training deep networks with unitary
  matrices.
\newblock In {\em Advances in Neural Information Processing Systems}.

\bibitem{ewc}
James Kirkpatrick, Razvan Pascanu, Neil Rabinowitz, Joel Veness, Guillaume
  Desjardins, Andrei~A Rusu, Kieran Milan, John Quan, Tiago Ramalho, Agnieszka
  Grabska-Barwinska, et~al.
\newblock Overcoming catastrophic forgetting in neural networks.
\newblock {\em Proceedings of the national academy of sciences},
  114(13):3521--3526, 2017.

\bibitem{kolesnikov2020big}
Alexander Kolesnikov, Lucas Beyer, Xiaohua Zhai, Joan Puigcerver, Jessica Yung,
  Sylvain Gelly, and Neil Houlsby.
\newblock Big transfer (bit): General visual representation learning.
\newblock In {\em Computer Vision--ECCV 2020: 16th European Conference,
  Glasgow, UK, August 23--28, 2020, Proceedings, Part V 16}, pages 491--507.
  Springer, 2020.

\bibitem{prompt_tuning}
Brian Lester, Rami Al-Rfou, and Noah Constant.
\newblock The power of scale for parameter-efficient prompt tuning.
\newblock In {\em Proceedings of the 2021 Conference on Empirical Methods in
  Natural Language Processing}, pages 3045--3059, Online and Punta Cana,
  Dominican Republic, November 2021. Association for Computational Linguistics.

\bibitem{bart}
Mike Lewis, Yinhan Liu, Naman Goyal, Marjan Ghazvininejad, Abdelrahman Mohamed,
  Omer Levy, Veselin Stoyanov, and Luke Zettlemoyer.
\newblock {BART:} denoising sequence-to-sequence pre-training for natural
  language generation, translation, and comprehension.
\newblock In Dan Jurafsky, Joyce Chai, Natalie Schluter, and Joel~R. Tetreault,
  editors, {\em Proceedings of the 58th Annual Meeting of the Association for
  Computational Linguistics, {ACL} 2020, Online, July 5-10, 2020}, pages
  7871--7880. Association for Computational Linguistics, 2020.

\bibitem{measureintrin}
Chunyuan Li, Heerad Farkhoor, Rosanne Liu, and Jason Yosinski.
\newblock Measuring the intrinsic dimension of objective landscapes.
\newblock In {\em International Conference on Learning Representations}, 2018.

\bibitem{ftgan2}
Qi~Li, Long Mai, Michael~A. Alcorn, and Anh Nguyen.
\newblock A cost-effective method for improving and re-purposing large,
  pre-trained gans by fine-tuning their class-embeddings.
\newblock {\em Asian Conference on Computer Vision}, 2020.

\bibitem{li2021prefix}
Xiang~Lisa Li and Percy Liang.
\newblock Prefix-tuning: Optimizing continuous prompts for generation.
\newblock In {\em Proceedings of the 59th Annual Meeting of the Association for
  Computational Linguistics and the 11th International Joint Conference on
  Natural Language Processing (Volume 1: Long Papers)}, pages 4582--4597, 2021.

\bibitem{li2020dylora}
Yinghui Li, Jing Yang, and Jiliang Wang.
\newblock Dylora: Towards energy efficient dynamic lora transmission control.
\newblock In {\em IEEE INFOCOM 2020-IEEE Conference on Computer
  Communications}, pages 2312--2320. IEEE, 2020.

\bibitem{liu2019roberta}
Yinhan Liu, Myle Ott, Naman Goyal, Jingfei Du, Mandar Joshi, Danqi Chen, Omer
  Levy, Mike Lewis, Luke Zettlemoyer, and Veselin Stoyanov.
\newblock Roberta: A robustly optimized bert pretraining approach.
\newblock {\em arXiv preprint arXiv:1907.11692}, 2019.

\bibitem{heavytailed}
Michael Mahoney and Charles Martin.
\newblock Traditional and heavy tailed self regularization in neural network
  models.
\newblock In {\em International Conference on Machine Learning}, pages
  4284--4293. PMLR, 2019.

\bibitem{matena2022merging}
Michael~S Matena and Colin~A Raffel.
\newblock Merging models with fisher-weighted averaging.
\newblock {\em Advances in Neural Information Processing Systems},
  35:17703--17716, 2022.

\bibitem{fedavg}
Brendan McMahan, Eider Moore, Daniel Ramage, Seth Hampson, and Blaise~Aguera
  y~Arcas.
\newblock Communication-efficient learning of deep networks from decentralized
  data.
\newblock In {\em Artificial intelligence and statistics}, pages 1273--1282.
  PMLR, 2017.

\bibitem{ftgan1}
Sangwoo Mo, Minsu Cho, and Jinwoo Shin.
\newblock Freeze the discriminator: a simple baseline for fine-tuning gans.
\newblock In {\em CVPR AI for Content Creation Workshop}, 2020.

\bibitem{vcl}
Cuong~V. Nguyen, Yingzhen Li, Thang~D. Bui, and Richard~E. Turner.
\newblock Variational continual learning.
\newblock In {\em International Conference on Learning Representations}, 2018.

\bibitem{ott2019fairseq}
Myle Ott, Sergey Edunov, Alexei Baevski, Angela Fan, Sam Gross, Nathan Ng,
  David Grangier, and Michael Auli.
\newblock fairseq: A fast, extensible toolkit for sequence modeling.
\newblock In {\em Proceedings of the 2019 Conference of the North American
  Chapter of the Association for Computational Linguistics (Demonstrations)},
  pages 48--53, 2019.

\bibitem{peftvsft2}
George Pu, Anirudh Jain, Jihan Yin, and Russell Kaplan.
\newblock Empirical analysis of the strengths and weaknesses of {PEFT}
  techniques for llms.
\newblock {\em CoRR}, abs/2304.14999, 2023.

\bibitem{LDM}
Robin Rombach, Andreas Blattmann, Dominik Lorenz, Patrick Esser, and Bj{\"o}rn
  Ommer.
\newblock High-resolution image synthesis with latent diffusion models.
\newblock In {\em Proceedings of the IEEE/CVF Conference on Computer Vision and
  Pattern Recognition}, pages 10684--10695, 2022.

\bibitem{ruiz2023dreambooth}
Nataniel Ruiz, Yuanzhen Li, Varun Jampani, Yael Pritch, Michael Rubinstein, and
  Kfir Aberman.
\newblock Dreambooth: Fine tuning text-to-image diffusion models for
  subject-driven generation.
\newblock In {\em Proceedings of the IEEE/CVF Conference on Computer Vision and
  Pattern Recognition}, 2023.

\bibitem{sharif2014cnn}
Ali Sharif~Razavian, Hossein Azizpour, Josephine Sullivan, and Stefan Carlsson.
\newblock Cnn features off-the-shelf: an astounding baseline for recognition.
\newblock In {\em Proceedings of the IEEE conference on computer vision and
  pattern recognition workshops}, pages 806--813, 2014.

\bibitem{lora_oninstructgpt}
Xianghui Sun, Yunjie Ji, Baochang Ma, and Xiangang Li.
\newblock A comparative study between full-parameter and lora-based fine-tuning
  on chinese instruction data for instruction following large language model.
\newblock {\em arXiv preprint arXiv:2304.08109}, 2023.

\bibitem{lpips}
Hossein Talebi and Peyman Milanfar.
\newblock Learned perceptual image enhancement.
\newblock In {\em 2018 IEEE international conference on computational
  photography (ICCP)}, pages 1--13. IEEE, 2018.

\bibitem{gradapprox1}
Thijs Vogels, Sai~Praneeth Karimireddy, and Martin Jaggi.
\newblock Powersgd: Practical low-rank gradient compression for distributed
  optimization.
\newblock In H.~Wallach, H.~Larochelle, A.~Beygelzimer, F.~d\textquotesingle
  Alch\'{e}-Buc, E.~Fox, and R.~Garnett, editors, {\em Advances in Neural
  Information Processing Systems}, volume~32. Curran Associates, Inc., 2019.

\bibitem{wang-etal-2018-glue}
Alex Wang, Amanpreet Singh, Julian Michael, Felix Hill, Omer Levy, and Samuel
  Bowman.
\newblock {GLUE}: A multi-task benchmark and analysis platform for natural
  language understanding.
\newblock In {\em Proceedings of the 2018 {EMNLP} Workshop {B}lackbox{NLP}:
  Analyzing and Interpreting Neural Networks for {NLP}}, pages 353--355,
  Brussels, Belgium, November 2018. Association for Computational Linguistics.

\bibitem{modelsoup}
Mitchell Wortsman, Gabriel Ilharco, Samir~Ya Gadre, Rebecca Roelofs, Raphael
  Gontijo-Lopes, Ari~S Morcos, Hongseok Namkoong, Ali Farhadi, Yair Carmon,
  Simon Kornblith, et~al.
\newblock Model soups: averaging weights of multiple fine-tuned models improves
  accuracy without increasing inference time.
\newblock In {\em International Conference on Machine Learning}, pages
  23965--23998. PMLR, 2022.

\bibitem{robustft}
Mitchell Wortsman, Gabriel Ilharco, Jong~Wook Kim, Mike Li, Simon Kornblith,
  Rebecca Roelofs, Raphael~Gontijo Lopes, Hannaneh Hajishirzi, Ali Farhadi,
  Hongseok Namkoong, et~al.
\newblock Robust fine-tuning of zero-shot models.
\newblock In {\em Proceedings of the IEEE/CVF Conference on Computer Vision and
  Pattern Recognition}, pages 7959--7971, 2022.

\bibitem{yao2022zeroquant}
Zhewei Yao, Reza Yazdani~Aminabadi, Minjia Zhang, Xiaoxia Wu, Conglong Li, and
  Yuxiong He.
\newblock Zeroquant: Efficient and affordable post-training quantization for
  large-scale transformers.
\newblock {\em Advances in Neural Information Processing Systems},
  35:27168--27183, 2022.

\bibitem{synint}
Friedemann Zenke, Ben Poole, and Surya Ganguli.
\newblock Continual learning through synaptic intelligence.
\newblock In {\em International conference on machine learning}, pages
  3987--3995. PMLR, 2017.

\bibitem{controlnet}
Lvmin Zhang and Maneesh Agrawala.
\newblock Adding conditional control to text-to-image diffusion models.
\newblock {\em arXiv preprint arXiv:2302.05543}, 2023.

\bibitem{zhangadaptive}
Qingru Zhang, Minshuo Chen, Alexander Bukharin, Pengcheng He, Yu~Cheng, Weizhu
  Chen, and Tuo Zhao.
\newblock Adaptive budget allocation for parameter-efficient fine-tuning.
\newblock In {\em The Eleventh International Conference on Learning
  Representations}.

\end{thebibliography}

\clearpage
% \counterwithin{figure}{section}
% \counterwithin{table}{section}

\newpage
\appendix
\onecolumn

\section{Detailed Explanation of the Budget Allocation Algorithm in ERE}
\label{sec:appendix_a}

In this section, we present a detailed algorithm for the budget-allocation  of our proposed ERE method. 
As described in \S\ref{sec:4.1}, our main objective is to solve the following optimization problem for rank variables $\textbf{r} \in \mathbb{Z}_{\geq 0}^N$
\begin{align*}
     \min_{\boldsymbol{r}} &\sum_{i=1}^{N} \sum_{l=r_i+1}^{\min(n_i, m_i)} \sigma_l^2 && \text{(Total Squared Error)} \\
    \text{subject to}\quad & \sum_{i=1}^{N} r_i(n_i+m_i) \leq M && \text{(Budget)}
\end{align*}
While total budget $M$ can be determined by the user, we find it convenient to set $M$ based on the prior rank $r$. This approach avoids the need to determine an appropriate budget $M$ for each model structure, which can be a cumbersome task. In our algorithm, denoted as ERE(r=$r_{avg}$) in our experiments \S\ref{sec:exp}, we set $M$ based on the input rank $r_{avg}$, such that $\sum_{i=1}^{N} r_{avg}(n_i+m_i) = M$. This definition allows us to interpret ERE as a process of reallocating ranks starting from a fixed-rank approximation of weight residuals.

Since finding an exact solution is generally NP-Complete, necessitating the use of approximate solutions. We address this challenge using a two-fold approach. First, we approximate the function $f_i(x) = \sum_{l = x + 1}^{\min(n_i, m_i)} \sigma_l^2$ using the spectral distribution to a suitable log-linear form, i.e., $g_i(x) = \exp{(a_ix + b_i)}$. This approximation allows us to relax the problem and solve it in a tractable manner using its simple derivative form.

Next, we utilize the approximated function $g_i$ to solve the equation using the method of Lagrangian multipliers. Assume that $M$ is small enough that the optimal solution lies at the constraint equality. We define the Lagrangian as follows:
\begin{equation*}
    L(\boldsymbol{r}, \lambda) = \sum_{i=1}^{N} g_i(r_i) + \lambda \left(\sum_{i=1}^{N} r_i (n_i + m_i) - M\right),
\end{equation*}
where $\lambda$ is the Lagrange multiplier. To find the optimal solution, we solve the following system of equations:
\begin{equation*}
    \frac{\partial L}{\partial r_i} = 0, \quad i=1,2,\ldots,N.
\end{equation*}
Taking the partial derivative of the Lagrangian with respect to $r_i$, we obtain:
\begin{equation*}
    \frac{\partial L}{\partial r_i} = g'_i(r_i) + \lambda (n_i + m_i) = 0.
\end{equation*}
Substituting this expression for $r_i$ into the constraint, we have:
\begin{equation*}
    \sum_{i=1}^{N} (g_i')^{-1}(-\lambda (n_i + m_i)) (n_i + m_i) = M.
\end{equation*}
In the special case of $g_i(x) =\exp(a_ix + b_i)$, we have $(g_i')^{-1}(x) = \frac{1}{a_i} \log{\frac{x}{a_i}} - \frac{b_i}{a_i}$. Thus, we obtain the following equation:
\begin{equation*}
\sum_{i=1}^{N} \frac{1}{a_i}\left(\log \left(-\frac{(n_i + m_i)}{a_i}\right) - b_i + \log \lambda\right) (n_i + m_i) = M.
\end{equation*}
While solving this equation for $\lambda$ and projecting $r$ back to the nearest solution provides a good solution, we can further improve it by considering the maximum rank condition imposed on $r_i$, i.e., $0 \leq r_i \leq \min(n_i, m_i)$. To incorporate this condition, we introduce the clamped variation by defining $u_i$ as follows:
\begin{equation*}
u_i(\lambda) = \text{clamp}\left(\frac{1}{a_i}\left(\log \left(-\frac{(n_i + m_i)}{a_i}\right) - b_i + \log \lambda\right), 0, \min(n_i, m_i)\right).
\end{equation*}
With the remaining variables held constant, we need to find an appropriate value of $\lambda$ such that $C(\lambda) = \sum_{i=1}^{N} u_i(\lambda) (n_i + m_i) = M$. This can be achieved through a binary search, as the function $C(\lambda)$ is monotonic with respect to $\lambda$. \autoref{alg:approx_solution}summarizes the procedure.

\begin{algorithm}[H]
\caption{ERE, budget allocation}\label{alg:approx_solution}
\SetKwInOut{Input}{Input}
\SetKwInOut{Output}{Output}
\SetAlgoLined
\Input{$r_{avg}$ (prior rank), $\Delta \theta_1, \Delta \theta_2, \ldots, \Delta \theta_N$ (residual weights)}
\Output{$\textbf{r}$ (approximate solution)}
Initialize $M = \sum_{i=1}^{N} r_{avg}(n_i+m_i)$
\For{$i=1$ \textbf{to} $N$}{
    Compute the singular values $\sigma_{i,1}, \sigma_{i,2}, \ldots, \sigma_{i,\min(n_i, m_i)}$ from  $\Delta \theta_i$;
    
    Compute $f_i(x) = \sum_{l = x + 1}^{\min(n_i, m_i)} \sigma_{i,l}^2$ for all $x \in [0, \min(n_i, m_i)]$\;
    Fit a log-linear model $g_i(r) = \exp(a_ir + b_i)$ to approximate $f_i(r)$\;
}

Initialize $\lambda_{\text{min}}, \lambda_{\text{max}}, \epsilon$ with suitable values for binary search\;

\While{$\lambda_{\text{max}} - \lambda_{\text{min}} > \epsilon$}{
    $\lambda \gets (\lambda_{\text{min}} + \lambda_{\text{max}}) / 2$\;
    Compute $C(\lambda) = \sum_{i=1}^{N} u_i(\lambda) (n_i + m_i)$\;
    \If{$C(\lambda) > M$}{
        $\lambda_{\text{max}} \gets \lambda$\;
    }
    \ElseIf{$C(\lambda) < M$}{
        $\lambda_{\text{min}} \gets \lambda$\;
    }
    \Else{
        \textbf{break}\;
    }
}
\For{$i=1$ \textbf{to} $N$}{
    $r_i \gets \text{clamp}\left(\frac{1}{a_i}\left(\log \left(-\frac{(n_i + m_i)}{a_i}\right) - b_i + \log \lambda\right), 0, \min(n_i, m_i)\right)$\;
    $r_i \gets \text{round}(r_i)$;
}
\textbf{return} $\textbf{r}$\;

\end{algorithm}

We would like to emphasize that our objective is based on the heuristic of minimizing the overall Frobenius norm, and the log-linear approximation of $f_i$ serves as a suitable proxy objective. However, we acknowledge that this metric function may not be the optimal choice and there might exist better schemes or more suitable solutions. Exploring alternative metrics and investigating improved approaches are left as future work to enhance the performance of the algorithm.

\section{Ablation Study: Evaluating Rank Allocation and Orthogonal Projection}
\label{sec:appendix_b}

In this section, we conduct an ablation study to evaluate the effectiveness of rank allocation and the influence of orthogonal projection. We build upon the settings described in \S\ref{effrank} and \S\ref{perturbable}, where we perform experiments on RoBERTa-Large models and analyze their feature outputs on 4 natural NLU tasks of GLUE benchmark.

\subsection{Ablation of Rank Allocation and Prior Rank}

First, we conduct an ablation study to examine the effectiveness of layer-wise rank allocation. We fix the prior rank ERE(r = 128) and vary the prior alpha $\alpha$ from 0 to 1 to achieve uniformization, as described in \S\ref{further_opt}. When $\alpha = 0$, it implies that we strictly follow the rank determined by the initial solution. 
On the other hand, setting $\alpha = 1$ is equivalent to setting equal rank $r_{avg}$ for all layers, which is identical to Low-Rank Approximation settings (LRA) in \S\ref{nlu_experiment}.

\begin{figure}[ht!]
  \centering
    \includegraphics[width=0.95\textwidth]{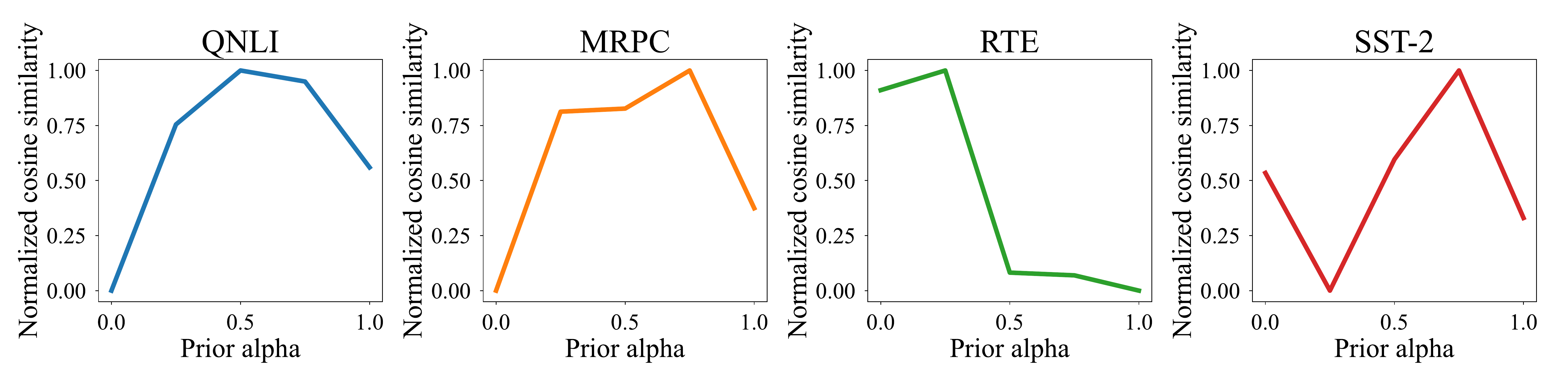}
    \caption{
    Effectiveness of prior-rank for different values of prior alpha, $\alpha$ in 4 GLUE tasks.
    }
    \label{fig:alpha_cos_sim}
\vspace{-3mm}
\end{figure}

From the analysis presented in \autoref{fig:alpha_cos_sim}, it is evident that the cosine similarity of the final layer output between the fine-tuned model and its ERE-approximated counterparts is best aligned for specific values of $\alpha \in (0, 1)$. This finding suggests that neither fully adhering to the solution nor setting all ranks equally yields optimal results.

In particular, when $\alpha = 0.0$, there is a possibility that some layers may have a rank of zero. This scenario can lead to a significant activation shift as one progresses through the network layers. To illustrate this phenomenon, let's consider a simplified example of an MLP with $N$ layers. Even if all layers have equal rank requirements, the initial layers inherently possess greater importance than the later layers. This is because errors originating in the early layers tend to propagate throughout the network, accumulating and magnifying in significance as they traverse the subsequent layers.

The imbalance in importance arises from the computational order of the layers, which is not explicitly taken into account in our algorithm. We treat each layer independently of others in terms of computational order, neglecting the inherent importance attributed to the earlier layers. Therefore, the effectiveness of our method is dependent on incorporating an appropriate value of alpha, as the objective tends to be misaligned, leading to extreme solutions that overly emphasize the significance of one layer compared to others.

However, the analysis presented in \autoref{fig:alpha_cos_sim} demonstrates that the promise of our algorithm remains intact. With the incorporation of an appropriate value of $\alpha$, our method consistently outperforms the uniform rank allocation approach.

\subsection{Stiefel Projection during Dequantization}

\begin{figure}[ht!]
  \centering
    \includegraphics[width=0.5\textwidth]{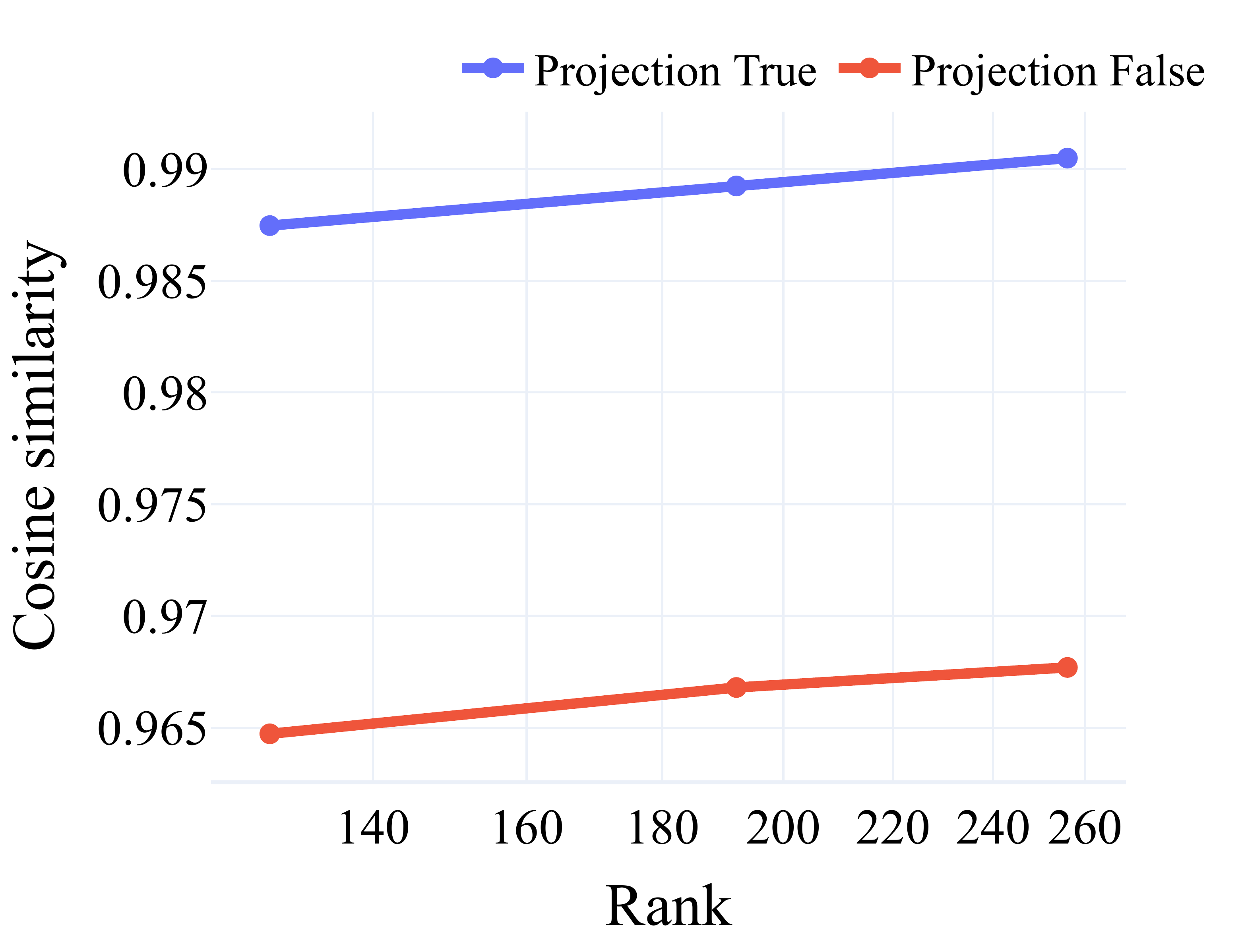}
    \caption{
    Effectiveness of the Stiefel projection. The blue line indicates applying projection during dequantization, and the red line indicates otherwise.
    }
    \label{fig:doroth}
\end{figure}

Furthermore, we perform an ablation study to evaluate the effectiveness of the Stiefel projection method. When employing round-to-nearest quantization on the $U$ and $V$ matrices, the use of small bit quantization may result in the loss of the orthogonal condition that was previously upheld by these matrices. To address this issue, we can perform a projection onto the nearest Stiefel manifold during the dequantization process. We leverage the following lemma for Stiefel projection \cite{unitary}.

\newtheorem*{lemma}{Lemma}
\begin{lemma}
Suppose that $X = UP$ is a polar decomposition of a matrix $X \in \mathbb{R}^{m\times n}$ ($m \ge n$). Then,
\begin{equation*}
\|X - U\| = \min\{\|X - V\| \mathrel{:} V \in \mathbb{S}^{m, n}\}
\end{equation*}
\end{lemma}

From the results presented in \autoref{fig:doroth}, we can clearly see the beneficial impact of the Stiefel projection. During the dequantization phase, applying the projection to the nearest Stiefel manifold significantly enhances the quality of weight residual reconstruction in terms of feature similarity.

\end{document}